\pdfoutput=1

\documentclass[11pt]{article}

\usepackage{listings}
\usepackage{booktabs}           
\usepackage{amsmath}            
\usepackage{graphicx,array}
\usepackage{multirow}
\usepackage{adjustbox}

\usepackage{acl}

\usepackage{times}
\usepackage{latexsym}

\usepackage[T1]{fontenc}

\usepackage[utf8]{inputenc}

\usepackage{microtype}

\usepackage{inconsolata}

\title{Multiple-Choice Questions are Efficient and Robust LLM Evaluators}

\author{
    Ziyin Zhang \quad Zhaokun Jiang \quad Lizhen Xu \quad Hongkun Hao \quad Rui Wang\thanks{Corresponding author.} \\
    Shanghai Jiao Tong University\\
    \texttt{\{daenerystargaryen, wangrui12\}@sjtu.edu.cn}
}

\begin{document}
\maketitle
\begin{abstract}
We present GSM-MC, a multiple-choice (MC) dataset constructed by collecting answers and incorrect predictions on GSM8K from 60 open-source models. Through extensive experiments, we show that LLMs' performance on the MC version of this popular benchmark is strongly correlated with their performance on the original version and is quite robust to distractor choices and option orders, while the evaluation time is reduced by a factor of up to 30. Following similar procedures, we introduce MATH-MC, constructed from MATH, and PythonIO, a new program reasoning MC dataset constructed from HumanEval and MBPP. Experimental results indicate that LLMs' performance on these MC benchmarks leaves much room for improvement.
Our data and code are available at \url{https://github.com/Geralt-Targaryen/MC-Evaluation}.
\end{abstract}

\section{Introduction}\label{sec:intro}

MMLU~\citep{2020MMLU}, GSM8K~\citep{2021GSM8K}, MATH~\citep{2021MATH}, HumanEval~\citep{2021HumanEval}, and MBPP~\citep{2021MBPP} are currently the de facto most popular benchmarks for evaluating LLMs. Among these benchmarks, only MMLU is in multiple-choice (MC) format, where model predictions can be efficiently extracted from output logits. In the other benchmarks, the models are typically evaluated by open-ended generation, from which the answers are extracted.

\begin{figure}[t]
    \centering
    \includegraphics[width=1\linewidth]{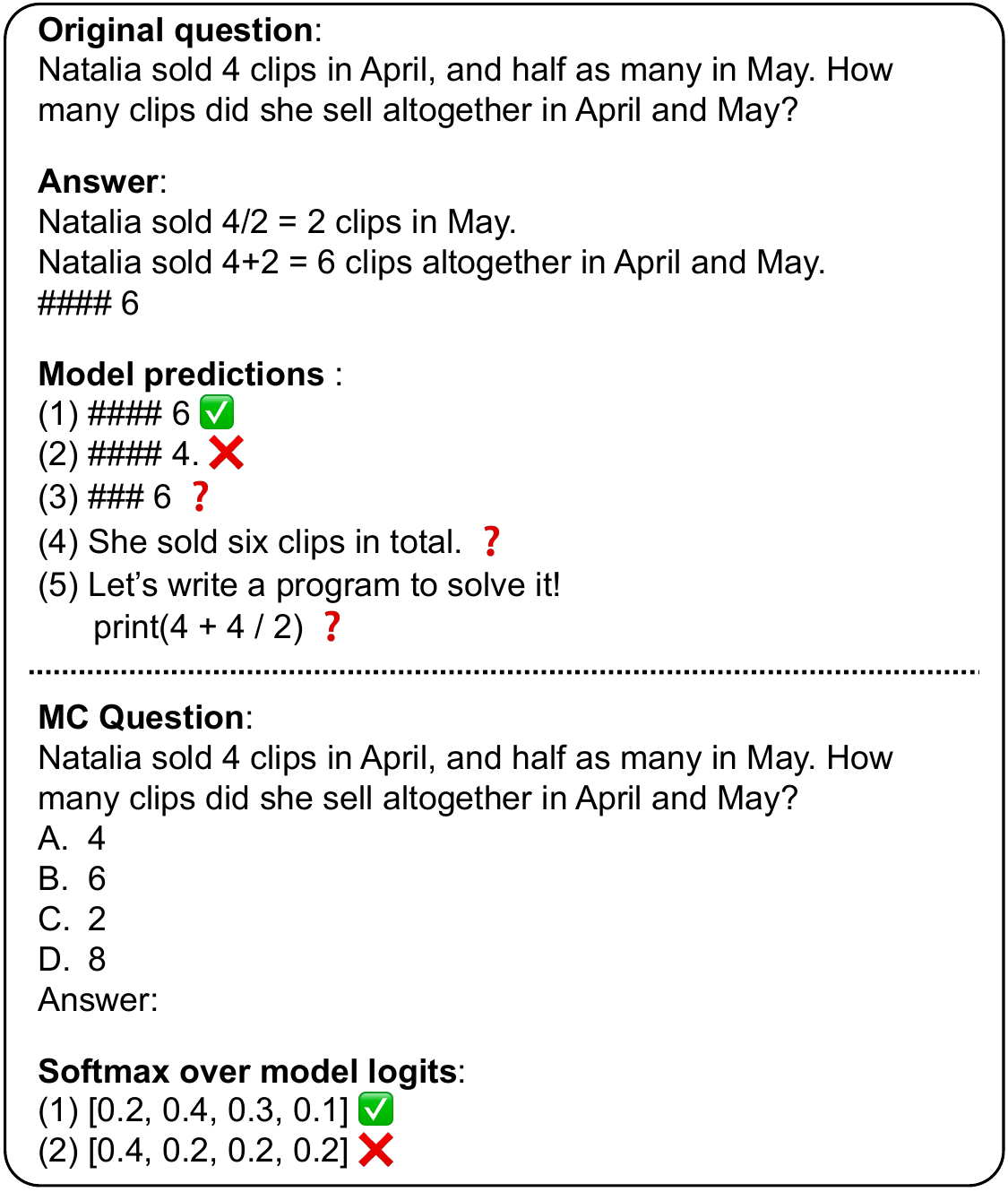}
    \caption{An illustrative example of correct, incorrect, and invalid answers to one question from GSM8K (top). After converting to multiple-choice format (bottom), a prediction can always be extracted from model logits.}
    \label{fig:demo}
\end{figure}

However, as shown in Figure~\ref{fig:demo}, LLMs may not always follow the required answer format during generation, which results in many false negatives when the answers are heuristically extracted from model generations and evaluated by exact match, as in GSM8K and MATH.

To tackle this issue, in this work we investigate whether short-answer generation benchmarks like GSM8K and MATH can be converted into a multiple-choice format to prevent invalid answers like those in Figure~\ref{fig:demo} from affecting the evaluation accuracy of LLMs. Using GSM8K as a proof-of-concept example, we collect incorrect predictions from 60 open-source models to construct a pool of distractors for each problem and convert the problems into an MC format similar to MMLU (which we dub GSM-MC). Through extensive experiments involving different numbers of choices~(Section~\ref{sec:gsmmc-choice-num}) and robustness against different distractor choices and option orders~(Section~\ref{sec:gsmmc-robustness}), we show that LLMs' performance on GSM-MC is robust to distractors and option orders, and strongly correlated with the performance on the original GSM8K regardless of choice numbers (ranging from 2 to 8).

\begin{table*}[th]
    \centering
    \scalebox{0.92}{
    \begin{tabular}{lrrrr}
    \toprule
        Benchmark & Training Samples & Test Samples & Source & Domain \\
    \midrule
        GSM-MC & 7468 & 1319 & GSM8K & grade school math word problems \\
        MATH-MC & 7278 & 4914 & MATH & high school math competitions\\
        PythonIO & 966 & 1684 & HumanEval, MBPP & Python program output prediction\\
    \bottomrule
    \end{tabular}
    }
    \caption{Overview of our MC datasets.}
    \label{tab:stats}
\end{table*}

Inspired by the success of converting GSM8K to GSM-MC, we repeat the same procedure on MATH to construct MATH-MC. The two coding benchmarks, however, can not be naively converted into MC format in the same way, which would result in outrageously long and very unnatural questions. Thus we follow one recent work~\citep{CRUXEval2024} and convert them into a program output prediction task instead, which we name PythonIO. An overview of these datasets is provided in Table~\ref{tab:stats}.
\section{Related Work}\label{sec:related}

The evaluation of LLMs can be categorized as either generation-based or multiple-choice-based. To compute a model's score on one specific generation sample - such as a math word problem in GSM8K~\citep{2021GSM8K} or one program synthesis problem in HumanEval~\citep{2021HumanEval} - there are several classes of metrics: 1) the first one is based on content overlap such as exact match, BLEU~\citep{2002BLEU}, and ROUGE~\citep{2004ROUGE}; 2) the second one is based on model-scoring, either by computing similarities between model representations (e.g. BERTScore~\citealp{2019BERTScore}), or by regression (e.g. BLEURT~\citealp{2020BLEURT}), or by directly asking a powerful LLM to grade the sample~\citep{2023ChatbotArena}; 3) the third one, specific to code, is based on functional correctness, where a generated program is run against a set of tests to verify their functions~\citep{2021HumanEval,2023codesurvey}. Most reasoning-heavy evaluations - such as math and coding benchmarks, where a small lexical difference in the answer could completely change its semantics - adopt the first and the third types of metrics. However, these metrics also require rigorous and possibly labor-intensive post-processing of model generations to work correctly, as exemplified in Figure~\ref{fig:demo}.

On the other hand, to evaluate a model on one MC question - such as one from MMLU~\citep{2020MMLU} - a binary score is typically computed by checking whether the model assigns the largest probability to the correct option id (e.g. ``A'') among all the option ids. While some early works used other methods such as concatenating the content of each option to the question and comparing their likelihood~\citep{2020GPT3}, it has been argued in the literature that such methods underperform compared with directly asking the model to output the answer id~\citep{MC2022}.

Recently Several works have studied the effectiveness and robustness of evaluating LLMs on MC benchmarks. \citet{MC2023-code} evaluated GPT models on questions from a Python programming course, finding the models to struggle with questions that require analysis and reasoning about the code, such as output prediction. \citet{MC2023} analyzed 20 LLMs' performance on MMLU and other MC benchmarks, finding that LLMs a priori assign higher probability to certain answer ids. \citet{MC2024} also investigated LLMs' performance on MMLU, finding them to be sensitive to the ordering of the four options in the question. However, all these works focused on existing MC benchmarks. To our knowledge, no work has explored the possibility of converting generation benchmarks into MC format.


\section{Converting GSM8K to Multiple-Choice Format}\label{sec:gsm-mc}
We first use GSM8K - which is relatively small in size and can be straightforwardly converted into an MC format - as a proof-of-concept example and validate the rationality of converting short-answer generation benchmarks into MC format.

\begin{figure*}[th]
    \centering
    \includegraphics[width=1\textwidth]{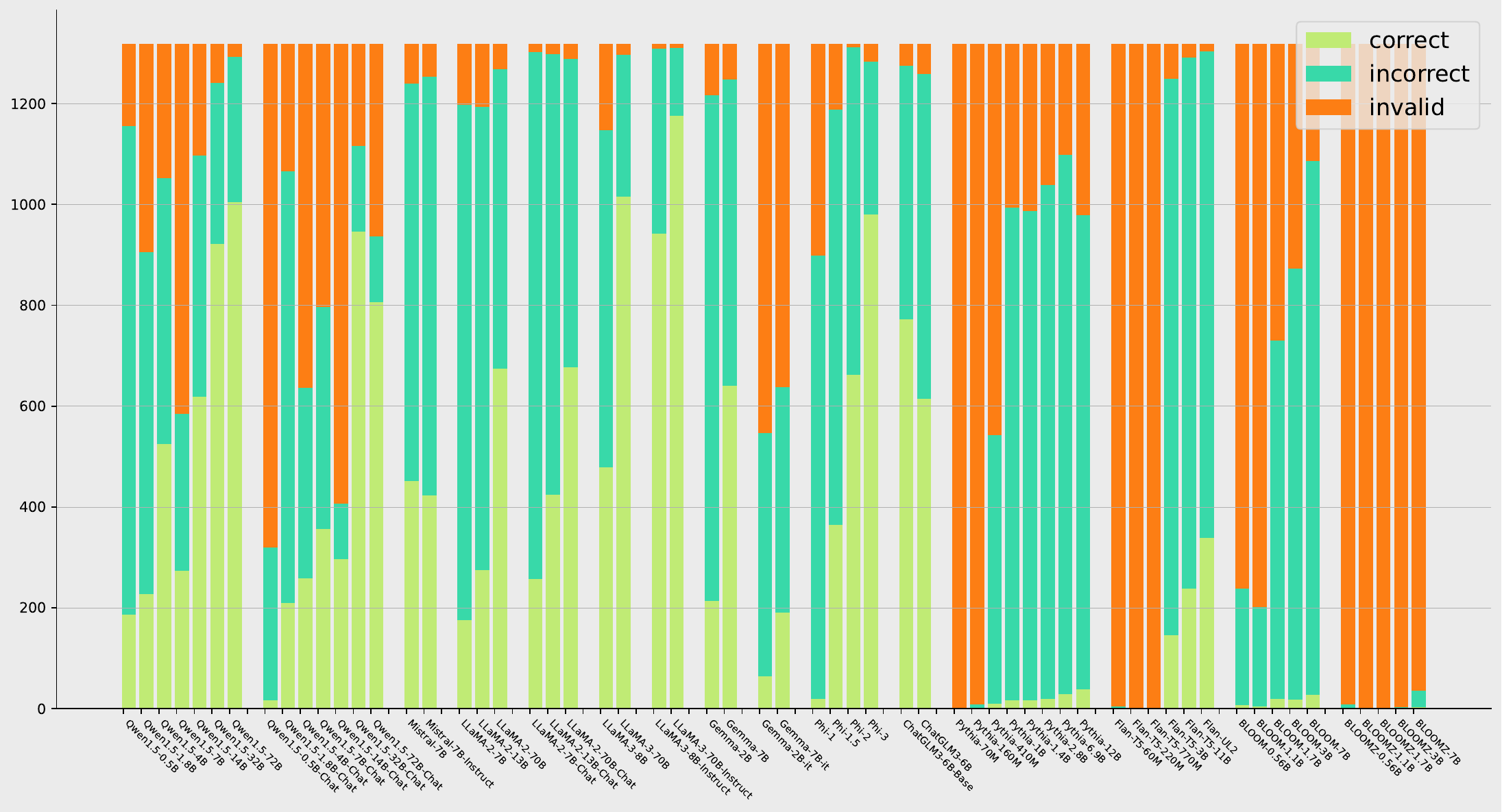}
    \caption{LLMs' answer distributions on GSM8K. Smaller models and aligned models tend to produce more invalid answers.}
    \label{fig:errors}
\end{figure*}

\begin{figure}[ht]
    \centering
    \includegraphics[width=1\linewidth]{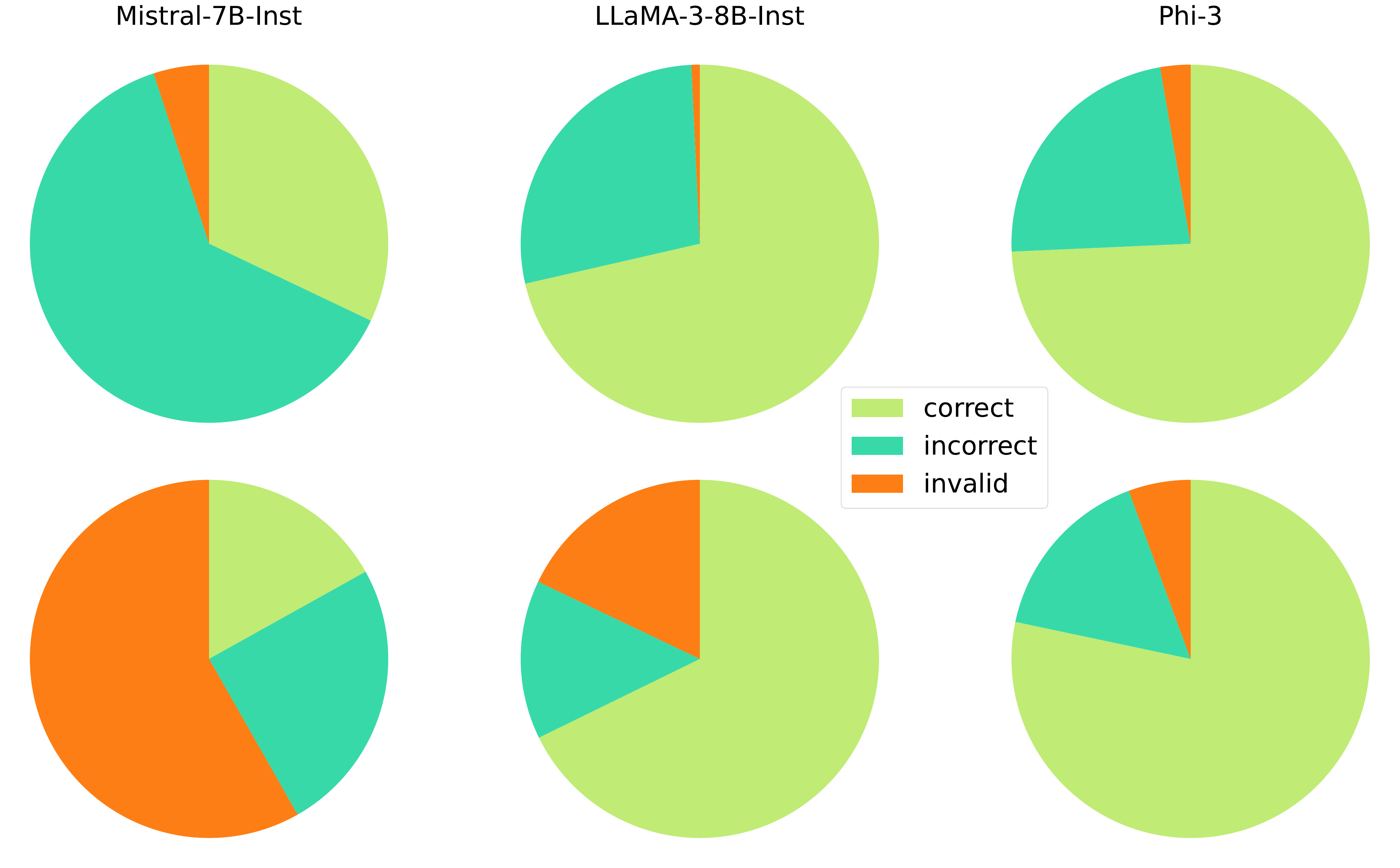}
    \caption{Comparison of answer distribution by aligned models with (top) and without (bottom) applying the instruction template.}
    \label{fig:chat-template}
\end{figure}

\subsection{A Closer Look at LLMs' Performance on GSM8K}\label{sec:gsmmc-problem}
Using the original prompt format provided by \citet{2021GSM8K}, we evaluated a series of open-source LLMs including Qwen1.5~\citep{2023Qwen}, LLaMA 2 and 3~\citep{2023LLaMA-2}, Mistral~\citep{2023Mistral}, Gemma~\citep{2024Gemma}, Phi 1-3~\citep{2023Phi-1,2023Phi-1.5,2024Phi-3}, ChatGLM3~\citep{2022GLM-130B}, Flan-T5~\citep{2019T5,2022Flan}, Pythia~\citep{2023Pythia}, and BLOOM~\citep{2022BLOOM,2022BLOOMZ} on GSM8K. The results indicate that a non-negligible portion of the wrong answers arises from failure to parse model outputs, as shown in Figure~\ref{fig:errors}. Inspecting the invalid answers, we identify three most common causes: meaningless repetition, not highlighting the answer in the correct format, and writing programs instead of solving the problems directly. More details about the prompt and sample outputs can be found in Figure~\ref{fig:prompt-orig},~\ref{fig:invalid-sample} in Appendix~\ref{sec:appendix-prompt}.

In the main experiments we used greedy decoding for all the evaluated models and did not apply any chat or instruction template for the aligned versions. In early experiments, We also evaluated the instruct versions of LLaMA-3, Mistral, and Phi-3 with their respective instruction template. As shown in Figure~\ref{fig:chat-template}, these templates lead to significantly more invalid responses, and also fewer correct answers for Mistral and LLaMA. We hypothesize that the instruction templates (for example, prepending \texttt{[INST]} and appending \texttt{[/INST]} to the prompt) interrupt the logical flow established by the consecutive in-context examples and make it harder for the models to follow the desired format.

\subsection{Converting to Multiple-Choice Format}\label{sec:gsmmc-convert}

To tackle the issue presented in Section~\ref{sec:gsmmc-problem}, we collected all the valid but incorrect answers produced by the evaluated models as a distractor pool for each problem in the benchmark. We then constructed a new dataset following the format of MMLU~(see Figure~\ref{fig:prompt-mc} in Appendix~\ref{sec:appendix-prompt}). We also additionally generated distractors for the training set to facilitate future research.

After converting to MC format, the evaluation process no longer involves auto-regressive generation but simplifies into one softmax operation over the option tokens' corresponding output logits at the end of the prompt. This leads to a significant reduction in computation cost: evaluating Qwen1.5-32B on the original GSM8K dataset takes 7 hours on our machine (distributed across 3 RTX 3090) while evaluating it on the newly constructed 4-way MC dataset takes only 13 minutes on the same machine.

\begin{figure*}[th]
    \centering
    \includegraphics[width=1\textwidth]{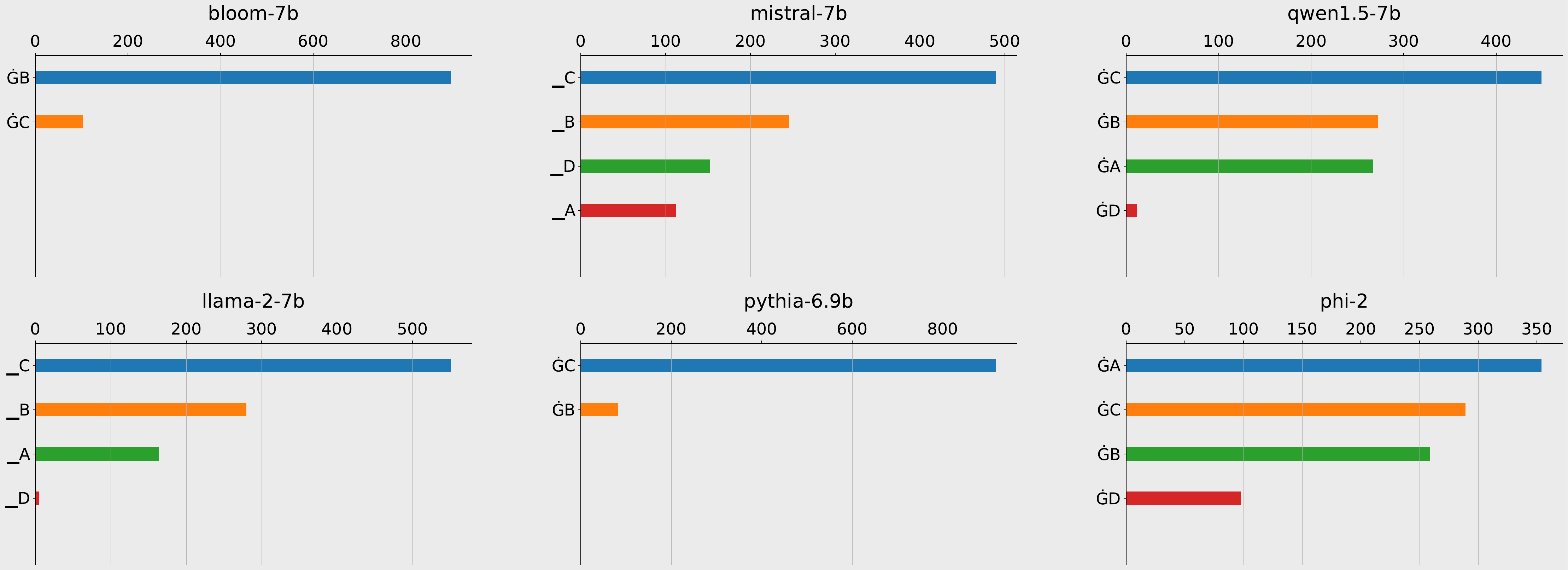}
    \includegraphics[width=1\textwidth]{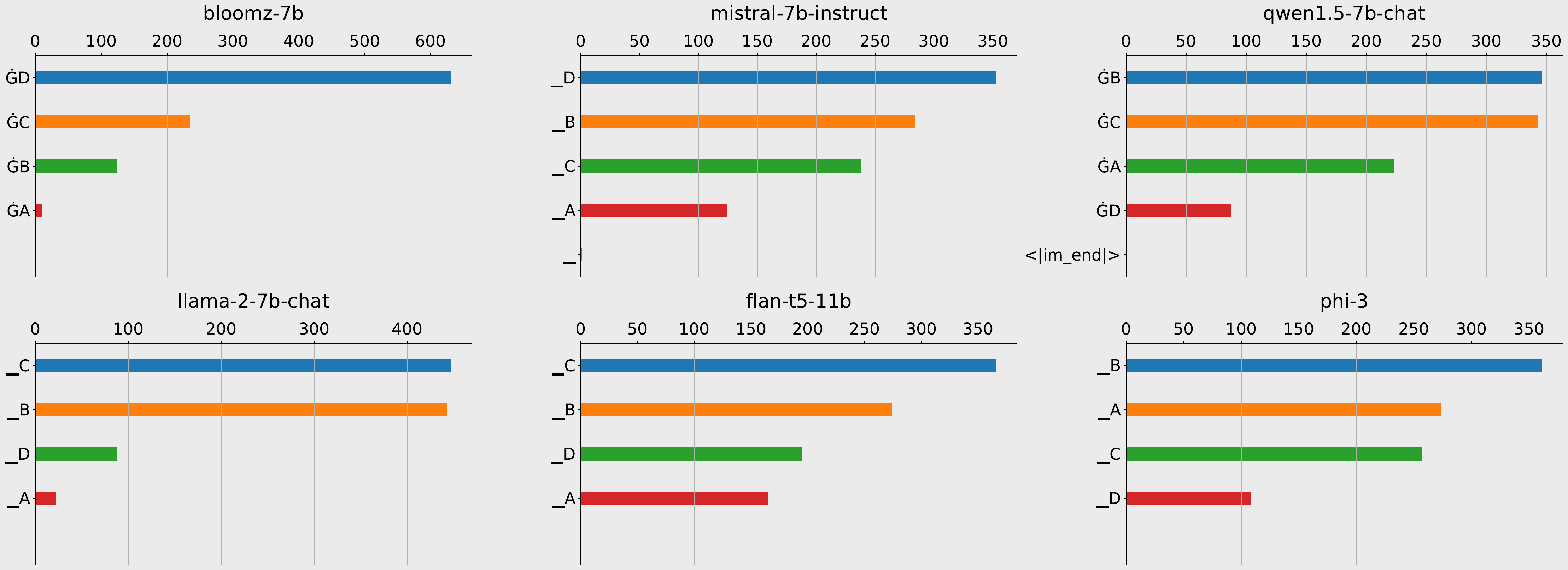}
    \caption{Frequency of most likely output token over 1K training set problems on GSM-MC by base models (top) and aligned models (bottom). The ground truth answers of the 1K problems are balanced across the four options.}
    \label{fig:mc-output-frequency}
\end{figure*}

\subsection{Can LLMs Understand Multiple-Choice Questions?}
As previously mentioned, one advantage of multiple-choice questions is that they enable the evaluation of any language model on any subject by simply comparing the output logits of the tokens ``A'', ``B'', ``C'', ``D''. However, the output logits of models are distributed over the entire vocabulary instead of only these option ids, and it remains unclear whether LLMs understand the multiple-choice format and tend to produce these tokens over other irrelevant tokens in the vocabulary. Thus, we first evaluated several models on one thousand 4-way MC problems constructed from the training set and counted the frequency of the most likely output token, presented in Figure~\ref{fig:mc-output-frequency}.

From the figure, we observe that \textbf{LLMs do understand multiple-choice format, but with a heavy bias towards certain options, which may be alleviated by alignment}. For example, both BLOOM 7B and Pythia 6.9B only outputs \texttt{B} and \texttt{C}, but never \texttt{A} and \texttt{D} for all the 1K problems, while the output distribution is more balanced in BLOOM's aligned version BLOOMZ.

Another issue Figure~\ref{fig:mc-output-frequency} reveals is the options tokenization. The currently most popular evaluation framework of MC problems provided by \citet{2020MMLU} directly tokenizes the options by calling \texttt{tokenizer("A").input\_ids[0].} However, this does not always yield the correct token id, since some tokenizers treat ``A'' and `` A'' as different tokens. For example, in LLaMA 3 tokenizer, the id of token ``A'' is 32, produced by the above script, while the id of token `` A'' is 362, one of the most likely tokens generated by the model after an MC question. In our implementation, we solve this issue by customizing the tokenization of options for each model.

\begin{figure*}[th]
    \centering
    \includegraphics[width=1\textwidth]{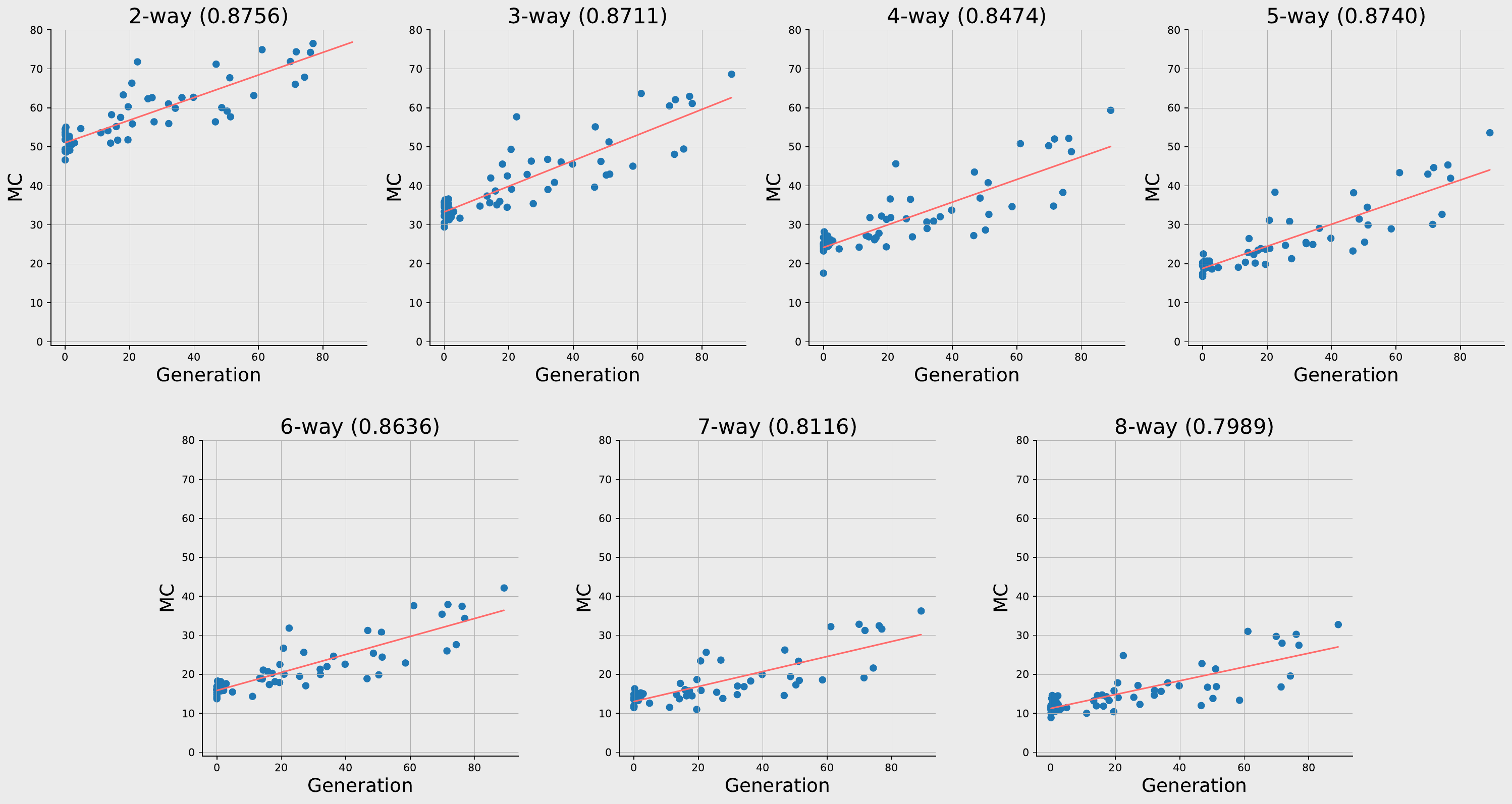}
    \caption{Model performance on GSM-MC (with the number of choices ranging from 2 to 8) and the original GSM8K. Each point is one model's score on GSM8K (x-axis) and one version of GSM-MC (y-axis), and the best-fitting line is given in red. The MC scores are strongly correlated with generation scores (Pearson correlation shown in each subplot's title), with a $p$-value less than 0.001 in all cases, indicating statistical significance.}
    \label{fig:generation-vs-mc}
\end{figure*}

\begin{figure*}[th]
    \centering
    \includegraphics[width=1\textwidth]{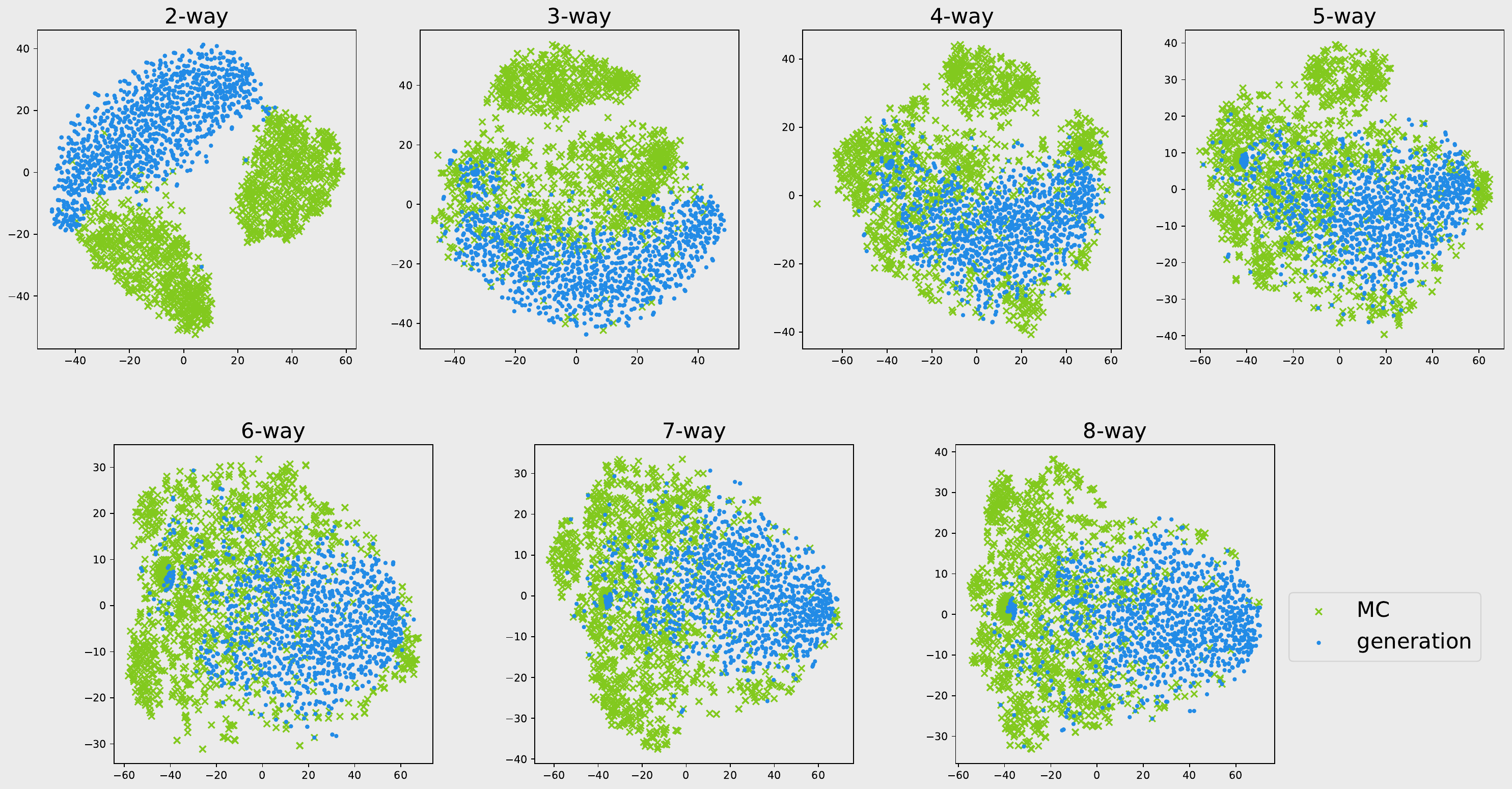}
    \caption{T-SNE visualization of questions in GSM8K and GSM-MC, using 40 models' correctness on each question as features. Starting from 4-way, the distribution of MC questions has a high overlap with generative questions.}
    \label{fig:tsne}
\end{figure*}

\subsection{Rationality of MC Evaluation}
\subsubsection{Correlation between MC Evaluation and Open-Ended Evaluation}\label{sec:gsmmc-choice-num}
From the experiments described in Section~\ref{sec:gsmmc-problem} and \ref{sec:gsmmc-convert}, we collected more than ten distractors for every problem in GSM8K's test set. Using these distractors, we constructed MC questions with different numbers of choices, ranging from 2-way to 8-way. We evaluated all the models mentioned in Section~\ref{sec:gsmmc-problem} on these seven suites of MC problems, and their performance is plotted against the performance on original GSM8K in Figure ~\ref{fig:generation-vs-mc}. \textbf{The results are strongly correlated with statistical significance in all cases.} 

To further explore the relation between models' performance on GSM8K and GSM-MC, we also visualize the questions in both datasets using the correctness of 40 models with non-trivial performance as features, as shown in Figure~\ref{fig:tsne}. In the 2-way setting, the MC questions are clearly structured into two clusters. This is explained by the fact that between the options ``A'' and ``B'', some models are biased towards the former while others are biased towards the latter (as shown in Figure~\ref{fig:mc-output-frequency}), which results in the features of questions with correct answer ``A'' and those with correct answer ``B'' having distinct distributions. However, as can be seen in Figure~\ref{fig:tsne}, \textbf{as the number of options increases in the MC questions, this correctness distribution gap is reduced, and the overall distribution of the MC questions also moves closer to that of the generative questions.}

Based on these findings, we consider 4-way MC problems by default in the rest of this work and in our released datasets, as 4-way questions are the most common MC format and our experiments also suggest that 4-way GSM-MC yields a similar model performance distribution to the original GSM8K. However, to contribute to future research, we also release all the distractors used to construct MC questions with more options.


\begin{figure*}[th]
    \centering
    \includegraphics[width=1\textwidth]{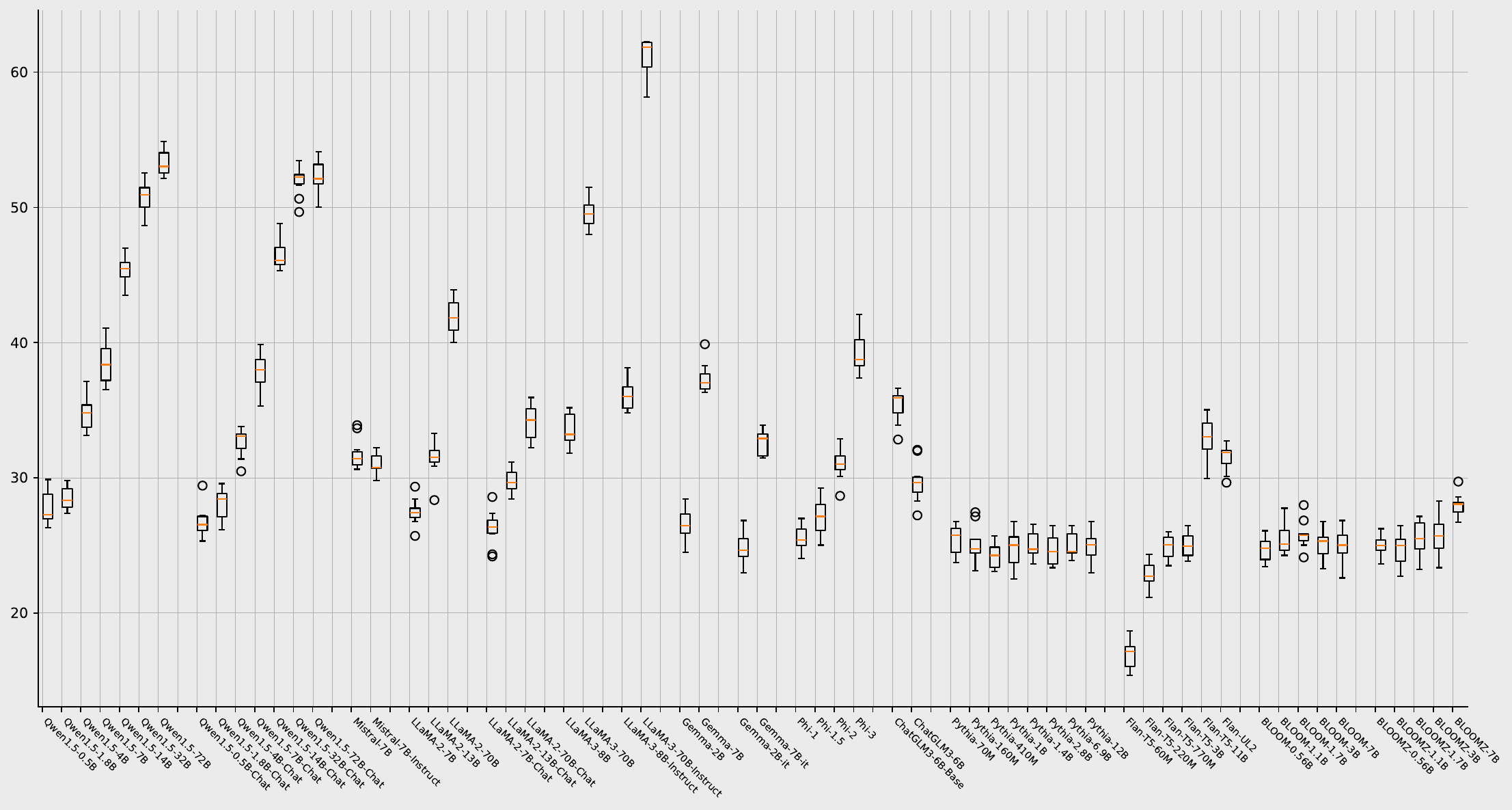}
    \caption{Performance variation on 4-way GSM-MC across ten sets of questions with different option orders and distractors.}
    \label{fig:performance-variation}
\end{figure*}

\subsubsection{Robustness against Distractors and Choice Orders}\label{sec:gsmmc-robustness}
Many works studying LLMs' performance on MC questions have suggested that LLMs are not robust to choice orders in MC problems~\citep{MC2022,MC2023,MC2024}. Thus, to explore LLMs' robustness on GSM-MC, we constructed ten different sets of 4-way MC problems from the distractor pool where both the choice of the three distractors and the order of the four options are randomized and repeated the previous experiments on these ten sets of problems. The results are plotted in Figure~\ref{fig:performance-variation}, where it can be observed that the variation of one model's performance is quite small compared with the inter-model difference.

\begin{table}[th]
    \centering
    \scalebox{0.94}{
    \begin{tabular}{lr>{\raggedleft\arraybackslash}m{3cm}}
    \toprule
        Distractors & Std & Correlation with generation scores \\
    \midrule
        model-generated & 1.083 & 0.859 \\
        randomly sampled & 1.017 & 0.705 \\
    \bottomrule
    \end{tabular}
    }
    \caption{Comparision of model scores on GSM-MC with model-generated and randomly sampled distractors: standard deviation across ten sets of questions, and mean correlation with the scores on the original GSM8K.}
    \label{tab:gen-vs-random}
\end{table}

We also experimented with an alternative strategy for generating distractors, where for a question with ground truth answer $n$, we randomly sample distractors in the interval $[0.5n, 1.5n]$\footnote{When $n$ is less than 40, we sample in $[n-20, n+20]$ instead.}. Like the previous experiment, we constructed ten sets of randomized questions and evaluated the models on them. We find that in this setting, the models' performance variation across the ten sets of problems is about the same as model-generated distractors, but the average correlation between scores on MC questions and the scores on the original GSM8K is much weaker, as shown in Table~\ref{tab:gen-vs-random}. Also, this strategy only applies to benchmarks where the ground truth answers are straightforward numbers, but fails in other cases (such as LaTeX expressions). Thus we recommend using model-generated distractors in future research.

\section{MATH-MC and PythonIO}\label{sec:math-python}
Inspired by the success of converting GSM8K to MC format, we also convert three other popular LLM evaluation benchmarks - MATH, HumanEval, and MBPP - into MC format to accelerate the evaluation of LLMs.

\begin{table*}[th]
    \small
    \centering
    \begin{tabular}{lrrrr}
    \toprule
        Model & GSM-MC & MATH-MC & PythonIO & Average \\
    \midrule
Qwen1.5-7B & 38.43$_{\pm1.43}$ & 42.96 & 32.78  & 38.06 \\
Qwen1.5-14B & 45.40$_{\pm0.92}$ & 50.65 & 40.86  & 45.64 \\
Qwen1.5-32B & 50.83$_{\pm1.10}$ & 54.48 & 51.78  & 52.36 \\
Qwen1.5-72B & 53.28$_{\pm0.89}$ & 55.92 & 50.36  & 53.19 \\
\\
Qwen1.5-7B-Chat & 37.85$_{\pm1.26}$ & 43.85 & 32.48  & 38.06 \\
Qwen1.5-14B-Chat & 46.46$_{\pm0.99}$ & 49.98 & 40.86  & 45.77 \\
Qwen1.5-32B-Chat & 51.92$_{\pm1.01}$ & 55.13 & 48.57  & 51.87 \\
Qwen1.5-72B-Chat & 52.30$_{\pm1.24}$ & 56.33 & 50.65  & 53.09 \\
\\
Mistral-7B & 31.74$_{\pm1.09}$ & 34.11 & 31.65  & 32.50 \\
Mistral-7B-Instruct & 31.00$_{\pm0.79}$ & 28.27 & 25.89  & 28.39 \\
\\
LLaMA-2-13B & 31.48$_{\pm1.27}$ & 30.12 & 26.60  & 29.40 \\
LLaMA-2-70B & 41.92$_{\pm1.22}$ & 40.64 & 38.24  & 40.27 \\
LLaMA-2-13B-Chat & 29.77$_{\pm0.79}$ & 28.94 & 28.03  & 28.91 \\
LLaMA-2-70B-Chat & 34.14$_{\pm1.27}$ & 32.36 & 31.47  & 32.66 \\
\\
LLaMA-3-8B & 33.52$_{\pm1.15}$ & 37.63 & 34.38  & 35.18 \\
LLaMA-3-70B & 49.58$_{\pm1.00}$ & 53.99 & 59.92  & 54.50 \\
LLaMA-3-8B-Instruct & 36.10$_{\pm1.07}$ & 37.61 & 38.95  & 37.55 \\
LLaMA-3-70B-Instruct & 61.14$_{\pm1.37}$ & 60.26 & 70.07  & 63.82 \\
\\
Gemma-7B & 37.33$_{\pm1.04}$ & 38.36 & 30.52  & 35.40 \\
Gemma-7B-it & 32.62$_{\pm0.90}$ & 33.52 & 27.97  & 31.37 \\
\\
Phi-2 & 30.98$_{\pm1.08}$ & 30.44 & 29.75  & 30.39 \\
Phi-3 & 39.26$_{\pm1.45}$ & 41.39 & 38.24  & 39.63 \\
\\
ChatGLM3-6B-Base & 35.32$_{\pm1.13}$ & 37.32 & 27.73  & 33.46 \\
ChatGLM3-6B & 29.69$_{\pm1.43}$ & 31.42 & 26.13  & 29.08 \\
\\
Flan-T5-3B & 25.03$_{\pm0.87}$ & 24.93 & 26.60  & 25.52 \\
Flan-T5-11B & 32.80$_{\pm1.63}$ & 26.43 & 29.33  & 29.52 \\
Flan-UL2 & 31.54$_{\pm0.97}$ & 27.35 & 25.95  & 28.28 \\
    \bottomrule
    \end{tabular}
    \caption{Selected results on GSM-MC, MATH-MC, and PythonIO. The results for GSM-MC are the mean value of the ten sets of different problems in Figure~\ref{fig:performance-variation}, with standard deviation given in subscript.}
    \label{tab:results}
\end{table*}

\paragraph{MATH} For MATH, we generated distractors with ChatGLM3, Qwen1.5, Gemma, Mistral, and Phi-2. As the answers are all latex expressions in this dataset, we used SymPy\footnote{\url{https://www.sympy.org/en/index.html}} to remove lexically different but semantically equivalent answers. After collecting the distractors, we filtered out a small number of questions where the ground truth answer extracted from the original solution is ambiguous (either empty or has more than one answer), which leaves us with 7.3K training samples and 4.9K test samples.

\paragraph{HumanEval and MBPP} For the code generation datasets, we follow \citet{CRUXEval2024} and convert the task into program output prediction instead. We heuristically extracted and manually verified input-output pairs from the unit tests in HumanEval and MBPP, and used Qwen1.5, Mistral, ChatGLM3, LLaMA-3, Phi-3, Gemma, and StarCoder~\citep{2023StarCoder} to generate distractors. We only retained distractors that can be successfully evaluated by a Python interpreter, and removed any duplicates. For the train/test split, we use all programs from HumanEval and the test set of MBPP as test samples, and the rest of MBPP as training samples.

The selected results of our evaluated models on MATH-MC and PythonIO, along with GSM-MC, are resented in Table~\ref{tab:results} (the complete results are given in Appendix~\ref{sec:appendix-results}). Overall, LLaMA-3 70B Instruct is the best performing model among all the evaluated models, scoring 61.1 on GSM-MC, 60.3 on MATH-MC, and 70.1 on PythonIO. Also, all three benchmarks prove to be rather challenging tasks, with few models scoring higher than 50, leaving much room for improvement.
\section{Conclusion}\label{sec:conclusion}

In this work, we convert two of the most popular LLM evaluation benchmarks - GSM8K and MATH - into multiple-choice format, and also construct a new program reasoning benchmark PythonIO from HumanEval and MBPP. Through extensive experiments, we show that LLMs' performance on GSM-MC is strongly correlated with their performance on the original GSM8K using open-ended generation, regardless of choice numbers and option orders. With the introduction of these three benchmarks, we hope to facilitate more efficient LLM evaluation in the research community.

\section*{Limitations}
Due to limited computation resources, throughout this work we used only GSM8K and GSM-MC as a proof-of-concept example to discuss the relation between a short-answer generation benchmark and its multiple-choice version. In terms of the other two benchmarks, MATH includes three times more questions than GSM8K, thus we expect most of the conclusions regarding robustness that we drew from experiments on GSM-MC to also hold on MATH-MC. As for PythonIO, the newly constructed benchmark evaluates a different capability (input-output reasoning) compared with the original HumanEval and MBPP (program synthesis), and is thus not directly comparable.

Also, the methodology taken in this work only applies to generation benchmarks with short, unique ground truth answers, but not other open-ended generation tasks such as machine translation and summarization. We leave the exploration of whether these tasks can also be evaluated more efficiently in multiple-choice format to future works.

\section*{Ethics Statement}
Regarding the data resources from which GSM-MC, MAHT-MC, and PythonIO are constructed, GSM8K, MATH, and HumanEval are released under MIT license, and MBPP is released under Apache 2.0 license. If you use, adapt, or redistribute our benchmarks, please also cite the original resources and include the corresponding license information. Our benchmarks should not be used outside of research contexts.

\bibliography{custom}

\onecolumn
\appendix

\section{Complete Results}\label{sec:appendix-results}
\begin{table*}[th]
    \small
    \centering
    \begin{tabular}{lrrrr}
    \toprule
        Model & GSM-MC & MATH-MC & PythonIO & Average \\
    \midrule
Qwen1.5-0.5B & 27.81$_{\pm1.21}$ & 25.11 & 25.18  & 26.03 \\
Qwen1.5-1.8B & 28.46$_{\pm0.80}$ & 28.90 & 26.43  & 27.93 \\
Qwen1.5-4B & 34.75$_{\pm1.15}$ & 37.81 & 25.71  & 32.76 \\
Qwen1.5-7B & 38.43$_{\pm1.43}$ & 42.96 & 32.78  & 38.06 \\
Qwen1.5-14B & 45.40$_{\pm0.92}$ & 50.65 & 40.86  & 45.64 \\
Qwen1.5-32B & 50.83$_{\pm1.10}$ & 54.48 & 51.78  & 52.36 \\
Qwen1.5-72B & 53.28$_{\pm0.89}$ & 55.92 & 50.36  & 53.19 \\
\\
Qwen1.5-0.5B-Chat & 26.75$_{\pm1.07}$ & 24.28 & 28.03  & 26.35 \\
Qwen1.5-1.8B-Chat & 28.08$_{\pm1.08}$ & 28.35 & 26.31  & 27.58 \\
Qwen1.5-4B-Chat & 32.68$_{\pm1.01}$ & 36.35 & 25.65  & 31.56 \\
Qwen1.5-7B-Chat & 37.85$_{\pm1.26}$ & 43.85 & 32.48  & 38.06 \\
Qwen1.5-14B-Chat & 46.46$_{\pm0.99}$ & 49.98 & 40.86  & 45.77 \\
Qwen1.5-32B-Chat & 51.92$_{\pm1.01}$ & 55.13 & 48.57  & 51.87 \\
Qwen1.5-72B-Chat & 52.30$_{\pm1.24}$ & 56.33 & 50.65  & 53.09 \\
\\
Mistral-7B & 31.74$_{\pm1.09}$ & 34.11 & 31.65  & 32.50 \\
Mistral-7B-Instruct & 31.00$_{\pm0.79}$ & 28.27 & 25.89  & 28.39 \\
\\
LLaMA-2-7B & 27.48$_{\pm0.92}$ & 29.08 & 23.04  & 26.53 \\
LLaMA-2-13B & 31.48$_{\pm1.27}$ & 30.12 & 26.60  & 29.40 \\
LLaMA-2-70B & 41.92$_{\pm1.22}$ & 40.64 & 38.24  & 40.27 \\
LLaMA-2-7B-Chat & 26.27$_{\pm1.25}$ & 26.48 & 25.53  & 26.09 \\
LLaMA-2-13B-Chat & 29.77$_{\pm0.79}$ & 28.94 & 28.03  & 28.91 \\
LLaMA-2-70B-Chat & 34.14$_{\pm1.27}$ & 32.36 & 31.47  & 32.66 \\
\\
LLaMA-3-8B & 33.52$_{\pm1.15}$ & 37.63 & 34.38  & 35.18 \\
LLaMA-3-70B & 49.58$_{\pm1.00}$ & 53.99 & 59.92  & 54.50 \\
LLaMA-3-8B-Instruct & 36.10$_{\pm1.07}$ & 37.61 & 38.95  & 37.55 \\
LLaMA-3-70B-Instruct & 61.14$_{\pm1.37}$ & 60.26 & 70.07  & 63.82 \\
\\
Gemma-2B & 26.50$_{\pm1.13}$ & 26.31 & 24.29  & 25.70 \\
Gemma-7B & 37.33$_{\pm1.04}$ & 38.36 & 30.52  & 35.40 \\
Gemma-2B-it & 24.82$_{\pm1.10}$ & 24.99 & 24.64  & 24.82 \\
Gemma-7B-it & 32.62$_{\pm0.90}$ & 33.52 & 27.97  & 31.37 \\
\\
Phi-1 & 25.46$_{\pm0.91}$ & 25.15 & 26.19  & 25.60 \\
Phi-1.5 & 27.09$_{\pm1.24}$ & 26.62 & 22.80  & 25.50 \\
Phi-2 & 30.98$_{\pm1.08}$ & 30.44 & 29.75  & 30.39 \\
Phi-3 & 39.26$_{\pm1.45}$ & 41.39 & 38.24  & 39.63 \\
\\
ChatGLM3-6B-Base & 35.32$_{\pm1.13}$ & 37.32 & 27.73  & 33.46 \\
ChatGLM3-6B & 29.69$_{\pm1.43}$ & 31.42 & 26.13  & 29.08 \\
    \bottomrule
    \end{tabular}
    \caption{The complete results on GSM-MC, MATH-MC, and PythonIO (continued in Table~\ref{tab:results-2}). The results for GSM-MC are the mean value of the ten sets of different problems in Figure~\ref{fig:performance-variation}, with standard deviation given in subscript.}
    \label{tab:results-1}
\end{table*}

\begin{table*}[th]
    \small
    \centering
    \begin{tabular}{lrrrr}
    \toprule
        Model & GSM-MC & MATH-MC & PythonIO & Average\\
    \midrule
Pythia-70M & 25.45$_{\pm1.03}$ & 26.21 & 27.08  & 26.25 \\
Pythia-160M & 25.05$_{\pm1.32}$ & 24.20 & 25.12  & 24.79 \\
Pythia-410M & 24.19$_{\pm0.88}$ & 24.93 & 23.69  & 24.27 \\
Pythia-1B & 24.64$_{\pm1.34}$ & 23.89 & 22.98  & 23.84 \\
Pythia-1.4B & 25.05$_{\pm0.88}$ & 24.60 & 23.28  & 24.31 \\
Pythia-2.8B & 24.63$_{\pm1.07}$ & 24.01 & 26.19  & 24.94 \\
Pythia-6.9B & 24.97$_{\pm0.92}$ & 23.54 & 25.12  & 24.54 \\
Pythia-12B & 24.93$_{\pm1.01}$ & 24.95 & 25.95  & 25.28 \\
\\
Flan-T5-60M & 16.95$_{\pm1.05}$ & 19.60 & 25.65  & 20.73 \\
Flan-T5-220M & 22.79$_{\pm0.99}$ & 22.59 & 24.05  & 23.14 \\
Flan-T5-770M & 24.93$_{\pm0.83}$ & 22.18 & 27.20  & 24.77 \\
Flan-T5-3B & 25.03$_{\pm0.87}$ & 24.93 & 26.60  & 25.52 \\
Flan-T5-11B & 32.80$_{\pm1.63}$ & 26.43 & 29.33  & 29.52 \\
Flan-UL2 & 31.54$_{\pm0.97}$ & 27.35 & 25.95  & 28.28 \\
\\
BLOOM-0.56B & 24.73$_{\pm0.86}$ & 23.97 & 25.42  & 24.71 \\
BLOOM-1.1B & 25.50$_{\pm1.16}$ & 24.97 & 24.17  & 24.88 \\
BLOOM-1.7B & 25.79$_{\pm0.98}$ & 25.23 & 23.16  & 24.73 \\
BLOOM-3B & 25.11$_{\pm1.01}$ & 25.17 & 24.35  & 24.88 \\
BLOOM-7B & 25.04$_{\pm1.19}$ & 25.03 & 24.23  & 24.77 \\
\\
BLOOMZ-0.56B & 25.05$_{\pm0.70}$ & 24.97 & 25.30  & 25.11 \\
BLOOMZ-1.1B & 24.76$_{\pm1.14}$ & 23.22 & 24.64  & 24.21 \\
BLOOMZ-1.7B & 25.48$_{\pm1.27}$ & 24.77 & 24.94  & 25.06 \\
BLOOMZ-3B & 25.71$_{\pm1.31}$ & 24.18 & 24.29  & 24.73 \\
BLOOMZ-7B & 27.94$_{\pm0.83}$ & 25.34 & 22.62  & 25.30 \\
    \bottomrule
    \end{tabular}
    \caption{The complete results on GSM-MC, MATH-MC, and PythonIO (continued from Table~\ref{tab:results-1}). The results for GSM-MC are the mean value of the ten sets of different problems in Figure~\ref{fig:performance-variation}, with standard deviation given in subscript.}
    \label{tab:results-2}
\end{table*}

\newpage
\phantom{.}
\newpage
\section{Prompt Details and Sample Outputs}\label{sec:appendix-prompt}

\lstset{
  basicstyle=\normalsize,
  columns=fullflexible,
  frame=single,
  breaklines=true,
  breakindent=0pt,
  moredelim=[is][\textbf]{\#\#\#\#\#\#}{\#\#\#\#\#\#},
}

\begin{figure*}[h]
\centering
\begin{lstlisting}
The following are grade school math word problems. Please answer the last problem following the format of the previous examples. Highlight the final answer with ####.

Natalia sold clips to 48 of her friends in April, and then she sold half as many clips in May. How many clips did Natalia sell altogether in April and May?
Natalia sold 48/2 = <<48/2=24>>24 clips in May.
Natalia sold 48+24 = <<48+24=72>>72 clips altogether in April and May.
#### 72

{four more in-context examples}

Janet's ducks lay 16 eggs per day. She eats three for breakfast every morning and bakes muffins for her friends every day with four. She sells the remainder at the farmers' market daily for $2 per fresh duck egg. How much in dollars does she make every day at the farmers' market?
\end{lstlisting}
\caption{Prompt format for evaluating LLMs on the original GSM8K dataset. In-context examples come from training set.}
\label{fig:prompt-orig}
\end{figure*}

\begin{figure*}[h]
\centering
\begin{lstlisting}
######Pythia 160M######
A $2.00 per day, a $1.00 a day, a $0.00 a day, a $0.00 a day, a $0.00 a day, a $0.00 a day, a $0.00 a day, a $0.00 a day, a $0.00 a day, a $0.00 a day, a $0.00 a day, a $0.00 a day, a $0.00 a day, a $0.00 a day, a $0.00 a day, a $0.00 a day, a $0.00 a day, a $0.00 a day, a $0.00 a day, ...

######BLOOM 1.7B######
How much does she make per day at the farmers' market?
Janet eats three eggs for breakfast every morning and bakes four muffins for her friends every day. She sells the remaining at the farmers' market daily for $2 per fresh duck egg. How much in dollars does she make every day at the farmers' market? How much does she make per day at the farmers' market?
Janet eats three eggs for breakfast every morning and bakes four muffins for her friends every day. She sells the remaining at the farmers' market daily for $2 per fresh duck egg. How much in dollars does she make every day at the farmers' market? How much does she make per day at the farmers' market?
...

######Flan-T5 770M######
Janet eats 3 eggs for breakfast and 4 for muffins every day for a total of 3 + 4 = 7 eggs. She sells the remaining 16 eggs at the farmers' market for $2 per egg for a total of 16 - 7 = $8 per day. The final answer: 8.

######Qwen1.5-Chat 14B######
Let's program in Python to solve it.
# define the variables
eggs_per_day = 16
breakfast_eggs = 3
muffin_eggs = 4
price_per_egg = 2

# calculate the eggs left after breakfast and baking
eggs_left = eggs_per_day - breakfast_eggs - muffin_eggs

# calculate the daily earnings
daily_earnings = eggs_left * price_per_egg

# print the result
print(daily_earnings)

\end{lstlisting}
\caption{Typical invalid responses to the question in Figure~\ref{fig:prompt-orig}: repetition at word-level (Pythia) and sentence-level (BLOOM), not following the answer format (Flan-T5), and writing programs instead of solving the problem directly (Qwen).}
\label{fig:invalid-sample}
\end{figure*}

\begin{figure*}[h]
\centering
\begin{lstlisting}
The following are multiple choice questions (with answers) about grade school math.

Natalia sold clips to 48 of her friends in April, and then she sold half as many clips in May. How many clips did Natalia sell altogether in April and May?
A. 30040
B. 84
C. 72
D. 96
Answer: C

{four more in-context examples}

Janet's ducks lay 16 eggs per day. She eats three for breakfast every morning and bakes muffins for her friends every day with four. She sells the remainder at the farmers' market daily for $2 per fresh duck egg. How much in dollars does she make every day at the farmers' market?
A. 22
B. 64
C. 18
D. 12
Answer:
\end{lstlisting}
\caption{Prompt format for evaluating LLMs on GSM8K-MC.}
\label{fig:prompt-mc}
\end{figure*}

\begin{figure*}[h]
\centering
\begin{lstlisting}
The following are multiple choice questions (with answers) about high school math.

A board game spinner is divided into three parts labeled $A$, $B$  and $C$. The probability of the spinner landing on $A$ is $\frac{1}{3}$ and the probability of the spinner landing on $B$ is $\frac{5}{12}$.  What is the probability of the spinner landing on $C$? Express your answer as a common fraction.
A. \frac{1}{12}
B. \dfrac{1-\frac{5}{12}}{12}
C. \frac{1}{4}
D. \frac{1}{1.67}
Answer: C

{four more in-context examples}

We roll a fair 6-sided die 5 times.  What is the probability that we get a 6 in at most 2 of the rolls?
A. \dfrac{50}{1296}
B. \frac{1}{4}
C. \frac{625}{648}
D. 1
Answer:
\end{lstlisting}
\caption{Prompt for evaluating LLMs on MATH-MC.}
\label{fig:prompt-mathmc}
\end{figure*}

\begin{figure*}[h]
\centering
\begin{lstlisting}
The following are multiple choice questions (with answers) about Python program reasoning.

Program:
R = 3
C = 3
def min_cost(cost, m, n): 
	tc = [[0 for x in range(C)] for x in range(R)] 
	tc[0][0] = cost[0][0] 
	for i in range(1, m+1): 
		tc[i][0] = tc[i-1][0] + cost[i][0] 
	for j in range(1, n+1): 
		tc[0][j] = tc[0][j-1] + cost[0][j] 
	for i in range(1, m+1): 
		for j in range(1, n+1): 
			tc[i][j] = min(tc[i-1][j-1], tc[i-1][j], tc[i][j-1]) + cost[i][j] 
	return tc[m][n]
Input:
min_cost([[1, 2, 3], [4, 8, 2], [1, 5, 3]], 2, 2)
Output:
A. 8
B. 10
C. 12
D. 6
Answer: A

{four more in-context examples}

Program:
def remove_Occ(s,ch): 
    for i in range(len(s)): 
        if (s[i] == ch): 
            s = s[0 : i] + s[i + 1:] 
            break
    for i in range(len(s) - 1,-1,-1):  
        if (s[i] == ch): 
            s = s[0 : i] + s[i + 1:] 
            break
    return s 
Input:
remove_Occ("hello","l")
Output:
A. "hell"
B. "heo"
C. "helo"
D. "hello"
Answer:
\end{lstlisting}
\caption{Prompt for evaluating LLMs on PythonIO.}
\label{fig:prompt-pythonio}
\end{figure*}

\end{document}